\documentclass[11pt]{article}

\usepackage[final]{acl}

\usepackage{times}
\usepackage{latexsym}
\usepackage{amsmath}
\usepackage{amssymb}
\usepackage{booktabs}
\usepackage{multirow} 
\usepackage{graphicx}
\usepackage{subcaption}
\usepackage{dblfloatfix}
\usepackage[T1]{fontenc}

\usepackage[utf8]{inputenc}

\usepackage{microtype}
\usepackage{afterpage}

\usepackage{inconsolata}

\usepackage{graphicx}

\title{Detect Before You Disambiguate: Perturbation-Based Input Ambiguity Detection in LLMs}

\author{
  \textbf{Tiejin Chen\textsuperscript{1,*}},
  \textbf{Xiaoou Liu\textsuperscript{1,*}},
  \textbf{Michael Clifford\textsuperscript{2}},
  \textbf{Ahmadreza Moradipari\textsuperscript{2}},
  \\
  \textbf{Yitao Chen\textsuperscript{2}},
  \textbf{Kyungtae Han\textsuperscript{2}},
  \textbf{Nejib Ammar\textsuperscript{2}},
  \textbf{Hua Wei\textsuperscript{1}}
  \\
  \\
  \textsuperscript{1}Arizona State University
  \qquad
  \textsuperscript{2}Toyota InfoTech Labs
  \\
  \small{\textsuperscript{*}Equal contribution.}
}

\begin{document}
\maketitle
\begin{abstract}
Large language models (LLMs) often answer ambiguous inputs as if they had a single intended meaning. This is problematic because input ambiguity is an upstream source of uncertainty: when a prompt admits multiple valid interpretations, the key question is whether the model should answer directly at all. Existing input uncertainty methods often rely on LLM-generated statements or clarifications, but such transformations may already disambiguate the input before the ambiguity is measured. We propose a label-preserving perturbation framework for input ambiguity detection in which perturbations remain close to the original input while preserving its label. We instantiate this principle with two detectors: a black-box rule-based perturbation library and a white-box importance-guided method using model-internal signals. Experiments across four datasets in open-domain, conversational, and visual QA show that our methods outperform output-side uncertainty baselines and LLM-paraphrase-based perturbation baselines. Our results highlight the importance of detecting ambiguity before attempting to disambiguate it. Our code is available \href{https://github.com/Xiao0o0o/input-ambiguity-detection}{\textcolor{purple}{here}}. 

\end{abstract}

\section{Introduction}
\label{sec:intro}

Large language models (LLMs) are increasingly deployed as question-answering systems, yet many real-world questions admit more than one legitimate interpretation~\cite{min-etal-2020-ambigqa,liu2023we, keyvan2022approach,liu2026mind}. When a question is ambiguous, a model cannot answer it well without first recognizing that an interpretive choice is required. After the ambiguity is recognized, a capable system has several reasonable options: it can ask the user a clarifying question, enumerate the plausible interpretations and answer each, or state the interpretation it is assuming \citep{zhang2025modeling, kuhn2022clam}. The problematic case is when the model silently commits to a single arbitrary interpretation and generates a single ``confident" answer, as if the question had no ambiguity at all. Consider a user who asks "\textit{When did the team win the championship?}" without specifying which team or which championship: a model that quietly settles on one reading may return a fluent, confident year that happens to answer a question the user never asked, with no recognition that other readings exist. Recent studies find that current models frequently fail at exactly this recognition step \citep{vijayvargiya2025interactive,zhang-etal-2024-clamber,kim2024aligning}.

Detecting whether a question is ambiguous is therefore a prerequisite for trustworthy question answering. If ambiguity can be detected, a system need not gamble on an arbitrary interpretation. 
We therefore frame the problem as a two-stage objective: first, \textit{detect} whether a question is ambiguous, and then, conditioned on that detection, \textit{mitigate} the ambiguity through targeted clarification or enumeration. Reliable detection is the foundation on which mitigation rests.

A natural and widely used strategy for detection is to probe the model's consistency: rephrase the question several ways, sample answers under each rephrasing, and treat disagreement among the answers as evidence of ambiguity \citep{hou2024decomposing, cole-etal-2023-selectively,kuhn2023semantic}. Intuitively, an unambiguous question should yield consistent answers while an ambiguous one should not. However, the existing implementation contains a subtle flaw: \textit{When a language model is used to rephrase the question, it must first understand the question, and understanding an ambiguous question entails resolving its ambiguity}. The rephrased question tends to encode a single interpretation chosen by the rephrasing model, eliminating the ambiguity that the procedure is meant to detect. 

Our core principle is that the step of ambiguity detection should not involve understanding the question. To preserve the ambiguity signal, the transformation applied to the input must be semantics-blind: it should stay close enough to the original input that it does not become a disambiguation step, yet remains expressive enough to expose interpretation-dependent behavior. We model this trade-off in the language of the information bottleneck \citep{tishby1999information}: we seek a perturbation that discards information about the input's surface form while preserving information about its ambiguity, and we use this principle to motivate our method rather than to strictly optimize it.

We instantiate this principle with two complementary detectors. (1) The white-box detector, which assumes access to model internals, uses internal signals to target a set of important input tokens before perturbing them. (2) The black-box detector, which assumes no access to model internals, applies a library of perturbation rules and measures the dispersion in the model's answer distribution.
Finally, we use the detection output to drive a lightweight re-asking mitigation strategy.

We evaluate on various datasets spanning open-domain, conversational, and visual question answering. We show that perturbation-based detection more reliably separates ambiguous from unambiguous questions than both output-side sampling and paraphrase-based baselines, with downstream gains when its output is used for mitigation.
\section{Related Work}

\paragraph{Ambiguity in LLMs}

Ambiguity is a natural property of user inputs because the same question may license several interpretations.  Prior work has studied this issue in open-domain and long-form QA through AmbigQA and ASQA~\cite{min-etal-2020-ambigqa,stelmakh-etal-2022-asqa}, and in interactive settings where ambiguity calls for clarification. Abg-CoQA annotates ambiguous conversational questions with clarification turns~\cite{guo2021abgcoqa}, while Abg-SciQA extends ambiguity analysis to scientific questions with labels for ambiguity detection, ambiguity type classification, clarification generation, and clarification-based QA~\cite{chen2025abgsciqa}. Other benchmarks cover proactive clarification in domain-specific QA~\cite{deng-etal-2022-pacific} and multimodal ambiguity in visual QA~\cite{jian-etal-2025-teaching,chen2025vqfocus,jang2026aqua}. Recent LLM-focused analyses further show that ambiguity is not solved simply by scale or prompting~\cite{chen2025abgsciqa}. LLMs can struggle to solve ambiguous problems~\cite {liu2023we,zhang-etal-2024-clamber,kamath-etal-2024-scope}. To solve the problem, prior work often studies what to do after ambiguity is recognized, including selectively abstaining from ambiguous questions~\cite{cole-etal-2023-selectively}, prompting models to detect, clarify, and answer~\cite{kuhn2022clam}, or building clarification-based QA pipelines such as CAmbigNQ~\cite{lee-etal-2023-asking}. Our work focuses on detecting whether the original input is ambiguous before resolution. This matters because resolution-oriented transformations, including generated paraphrases, answer sets, or clarification candidates, can potentially choose one interpretation from the ambiguity and remove the ambiguity being measured. 

\begin{figure*}
    \centering
    \includegraphics[width=0.9\linewidth]{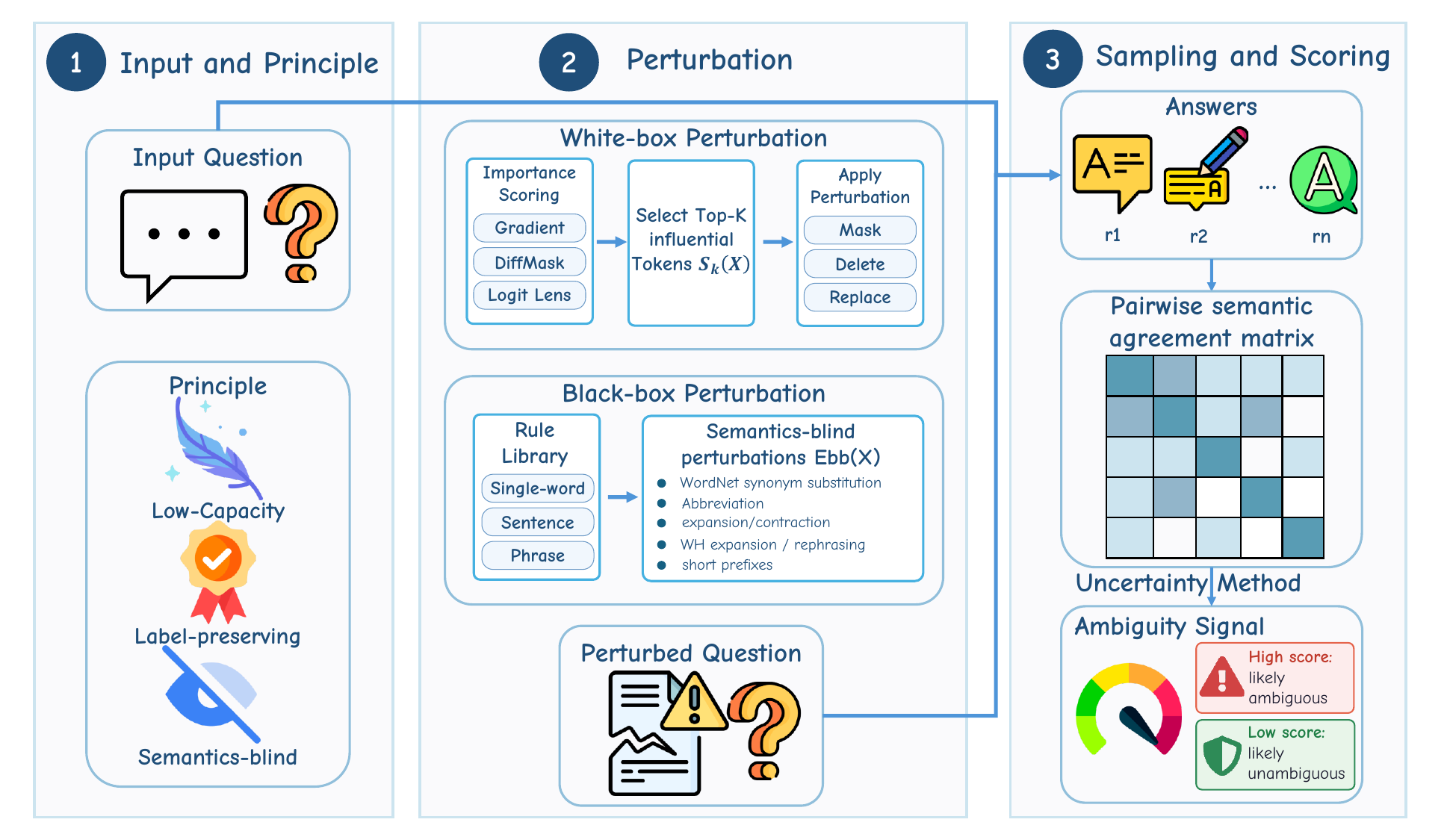}
    \caption{Overview of the proposed ambiguity detection framework. Given an input question and perturbation principles, the method generates label-preserving, low-capacity, and semantics-blind perturbations via white-box token-importance methods and black-box rule-based transformations. The perturbed questions are then sampled to obtain multiple model answers, whose pairwise semantic agreement is used to compute an ambiguity signal: low agreement indicates likely ambiguity, while high agreement suggests the question is likely unambiguous.}
    \vspace{-3mm}
    \label{fig:framework}
\end{figure*}

\paragraph{Uncertainty Quantification in LLMs}
Uncertainty Quantification (UQ) is considered important in trustworthy deployment of LLMs~\cite{liu2025uncertainty}. Existing UQ methods often estimate reliability from token probabilities, verbalized~\cite{kadavath2022language,lin2022teaching,ren2023selfevaluation,azaria2023internal,chen2024inside}, but the dominant black-box paradigm for free-form generation is multiple sampling, where several generations are drawn and compared~\cite{chen2026every,duan2024shifting, lin-etal-2024-contextualized, nikitin2024kernel,kuhn2023semantic,farquhar2024detecting}. For example, \citet{lin2024generating} treats uncertainty as dispersion over possible predictions for a fixed input, and proposes semantic-dispersion measures based on pairwise relations among sampled outputs.  Such output-side methods are natural baselines, but diverse generations can reflect either model unreliability or multiple valid interpretations of an ambiguous input. Input-side UQ~\cite{liu2025uncertainty} is therefore closer to our setting, and existing methods probe it mainly through prompt perturbations~\cite{hou2024decomposing,gao-etal-2024-spuq,yang2024just}. However, these transformations may remove ambiguity before it is measured. Our framework instead requires label-preserving perturbations, instantiated by a black-box or white-box method, so that measured variance better reflects input ambiguity rather than disambiguation introduced by the perturbation process.


\section{Methodology}
\label{sec:Method}

\subsection{Overall Framework}
\label{ssec:overall_framework}

\paragraph{Theoretical motivation.}
Pure output-side sampling can reveal disagreement among generated answers, but it cannot tell whether the disagreement comes from model unreliability or from multiple valid readings of the input. Direct paraphrasing or clarification generation probes the input side, but the transformation is high-capacity and can resolve the ambiguity before it is measured. To address both issues, our detector asks whether small, label-preserving interventions can expose answer instability without becoming a disambiguation step. This view is inspired by the information bottleneck objective~\cite{tishby1999information}:
\begin{equation}
\min_{p(t\mid x)} I(X;T)-\beta I(T;Y),
\end{equation}
where \(I(X;T)\) penalizes retaining too much information from the input, \(I(T;Y)\) rewards preserving information about the ambiguity label, and \(\beta\) controls this trade-off. We do not optimize a strict IB objective or estimate mutual information. Instead, the bottleneck is imposed by constraining the intervention channel.

\paragraph{Input and perturbation principle.}
As illustrated in Figure~\ref{fig:framework}, our pipeline starts from an input question and applies valid perturbations under three principles.
Let \(X\) be the original question, \(Y\in\{0,1\}\) the ambiguity label, and \(T=E(X)\) a perturbed question. A valid intervention \(E\) should satisfy three principles. \textbf{Low capacity.} The perturbation should make only a small surface-level change, so that \(T\) remains close to \(X\). \textbf{Label-preserving.} The perturbation should preserve the ambiguity label, so that \(T\) is ambiguous if and only if \(X\) is ambiguous. \textbf{Semantic-blind.} The perturbation may use lexical rules, surface templates, or importance scores to decide where to intervene, but it should not generate a semantic rewrite or choose among possible interpretations of the question. Together, these principles prevent the perturbation from turning ambiguity detection into ambiguity resolution by perturbation.

\paragraph{Perturbation pipeline.}
The perturbation module produces a small set of perturbed questions \(\mathcal{T}_k(X)=\{T_1,\ldots,T_k\}\). In the white-box setting, model-internal signals rank input tokens or short spans, and interventions such as replacement or deletion are applied to the top-ranked positions. In the black-box setting, a constrained rule library produces surface-level perturbations without using model internals. The top-\(k\) budget controls the number of perturbation candidates used for scoring rather than defining a separate maximization objective. We describe the white-box and black-box perturbation modules in the next two subsections.

\paragraph{Multiple sampling and scoring.}
After perturbation, we sample from both the original and perturbed questions. Specifically, we draw \(m=5\) answers from \(X\) and \(m=1\) answers from each \(T_j\in\mathcal{T}_k(X)\), then combine all sampled answers into the full answer set:
\begin{equation}
R_X^{\mathcal{T}}
=
\{r_i\}_{i=1}^{5}
\cup
\{\tilde r_j\}_{j=1}^{k},
\label{eq:answer_set}
\end{equation}

The resulting answer set can be scored by different uncertainty estimators. In this work, we use \(\mathrm{EigV}(\mathrm{Agree})\) from ~\citet{lin2024generating}. And the detail technology of \(\mathrm{EigV}(\mathrm{Agree})\) can be found in Appendix~\ref{sec:appendx_tech_uq}.

\subsection{White-Box Perturbation}
\label{ssec:whitebox_perturbation}

White-box perturbation uses model access to propose low-capacity interventions that are likely to produce informative answer distortion. We focus on token because sentence-level rewrites are too high capacity and can blur detection with resolution, while token-level changes give the cleanest bottleneck test. The white-box component is not used to rewrite the question. It only identifies where a semantics-blind intervention should be applied. Attention-based scores were less reliable in preliminary experiments, so we use three attribution methods that more directly track output sensitivity. Given token scores \(a_1,\ldots,a_n\), the intervention set is selected as
\begin{equation}
S_k(X)=\mathrm{TopK}_{i\in[n]} a_i,\qquad E_{\mathrm{wb}}(X)=E(X;S_k),
\end{equation}
where \(E(X;S_k)\) masks, deletes, or replaces only the selected tokens. The importance score only proposes \(E\), while ambiguity is measured by the multiple-sampling distortion described above.

\paragraph{Vanilla Gradient.}
Vanilla Gradient decodes the first token greedily, \(y^\star=\arg\max_y p_f(y\mid X)\), backpropagates \(\log p_f(y^\star\mid X)\) to the input embeddings, and scores each token by \(a_i=\|\nabla_{e_i}\log p_f(y^\star\mid X)\|_2\).

\paragraph{DiffMask.}
DiffMask learns a continuous mask over input embeddings~\cite{de-cao-etal-2020-decisions}. We adapt it to find a sparse set of tokens whose masking changes the output entropy,
\begin{equation}
\begin{aligned}
m^\star
&=\arg\max_{m\in[0,1]^n}
\left|H(p_f(\cdot\mid X_m))-H(p_f(\cdot\mid X))\right| \\
&\quad -\lambda\|m\|_1 .
\end{aligned}
\end{equation}
Here \(m_i\) controls how much token \(i\) is masked, \(X_m\) is the masked input, \(H(\cdot)\) is the entropy of the next-token distribution, and \(\lambda\) penalizes large masks so that the selected intervention remains low capacity.

\paragraph{Logit Lens.}
Logit Lens projects intermediate hidden states through the final layer normalization and unembedding matrix to obtain layerwise vocabulary distributions~\cite{nostalgebraist2020logitlens,belrose2023tunedlens}. We use these latent predictions to identify tokens whose internal trajectory becomes strongly aligned with the final answer or with answer-relevant alternatives. These tokens are plausible places where the model commits to one interpretation, making them useful targets for low-capacity intervention.

\subsection{Black-Box Perturbation}
\label{ssec:blackbox_perturbation}

The black-box setting assumes only sample-level query access to the model. Gradients, hidden states, and token-level scores are all unavailable, so the detector can observe nothing beyond the text the model returns. Under this constraint we cannot inspect where the model allocates attention or how confident it is internally; we can only perturb the input and watch how the output distribution responds. We therefore probe the input through a fixed library of \emph{semantics-blind} rewrites and identify informative rules as those that induce a large shift in the multi-sample output distribution. By semantics-blind we mean that each rule fires purely on surface features, namely lexical category, orthographic form, or question template, and never consults a model generated reading of the input. This property is what keeps detection separate from mitigation: a rule that does not understand the question cannot quietly resolve its ambiguity before we have had a chance to measure it.

\paragraph{Single-word rules.}
At the single-word level, we apply part-of-speech-constrained synonym substitutions for verbs, nouns, adjectives, and adverbs using WordNet~\cite{miller1995wordnet}. For each targeted word, we restrict substitution to synonyms drawn from its most frequent synset, filter out multi-word and rare candidates, and reinflect the chosen synonym so that its tense, number, or degree matches the original word form. This keeps each rewrite grammatical while changing only the lexical choice. We additionally include two orthographic rules that operate on the abbreviated forms of function words, namely contraction expansion and its inverse, contraction (for example \emph{Who's} $\leftrightarrow$ \emph{Who is}). Each of these rules alters a single content word or its surface form while leaving the referential structure of the question, and therefore the set of interpretations it admits, intact.

\paragraph{Phrase- and sentence-level rules.}
At the phrase level, we prepend short prefixes of three registers: test-like (for example \emph{According to the passage}), conversation-like (for example \emph{Hey so}), and instruction-like (for example \emph{Given the information provided}). These prefixes preserve the question stem verbatim and shift only its register, leaving the underlying query unchanged. We further apply two transformations to the leading wh word. WH expansion replaces a bare wh word with an equivalent determiner phrase (for example \emph{Who} $\rightarrow$ \emph{Which person}, \emph{When} $\rightarrow$ \emph{At what time}), and WH rephrasing substitutes the wh word with a semantically aligned alternative (for example \emph{Who} $\rightarrow$ \emph{What person}, \emph{How} $\rightarrow$ \emph{By what means}). Both wh rules are guarded so that they do not fire when the wh word is acting as a determiner, as in \emph{What color was it}, where expansion would distort the question; in such cases, only the conservative \emph{what}/\emph{which} swap is permitted. These edits are more disruptive than single-word rewrites but still draw on fixed templates rather than free generation, which keeps them within the semantics-blind regime.

\paragraph{Aggregating the ambiguity signal.}
Given a question $X$, we apply every applicable rule to obtain a set of variants $\{E_1(X), \dots, E_K(X)\}$, then draw $N$ samples from the model for the original question and for each variant under the same QA prompt and context. Rather than scoring each rewrite in isolation, we pool the model answers from the original question and all of its variants into a single set and compute one dispersion score over that pooled set using $\mathrm{EigV}(\mathrm{Agree})$. The intuition is that an unambiguous question yields answers that remain stable across the entire family of surface variants, whereas an ambiguous one fractures into competing answers as different rewrites nudge the model toward different readings. Aggregating across variants rather than scoring them separately gives a single, lower variance estimate of how much the answer distribution moves under semantics-preserving perturbation, which is precisely the quantity our information bottleneck view identifies with input ambiguity.




\section{Experiments}
Our evaluation addresses three research questions:

\noindent$\bullet$~\textbf{RQ1} (\S\ref{ssec:rq1}). Does our perturbation-based detector separate ambiguous from unambiguous questions more reliably than existing output-side and prompt-based baselines?

\noindent$\bullet$~\textbf{RQ2} (\S\ref{ssec:rq2}). What is the relationship between different perturbations and different importance? When applied in isolation, does each single perturbation type already carry an ambiguity signal, and how consistent are different attribution methods in terms of which words they flag?

\noindent$\bullet$~\textbf{RQ3} (\S\ref{ssec:rq3}). When detector output drives downstream mitigation, does it yield higher answer correctness than generation without detection?

\begin{table*}[t]
\centering
\small
\setlength{\tabcolsep}{5pt}
\renewcommand{\arraystretch}{1.05}
\caption{Ambiguity-detection AUROC across three model families and four QA benchmarks. \textbf{Bold} marks the best score within each model block per column, and \underline{underline} marks the second best. Vision columns use the vision-language counterpart of the text model (Qwen2.5-VL-3B for Qwen, InternVL3-8B for the Llama row, and the same Gemma-3-27B-it checkpoint for Gemma).}
\label{tab:main}
\begin{tabular}{l l ccccc}
\toprule
& & \multicolumn{2}{c}{\textbf{Text QA}} & \multicolumn{2}{c}{\textbf{Vision QA}} & \\
\cmidrule(lr){3-4}\cmidrule(lr){5-6}
\textbf{Model} & \textbf{Method} & \textbf{ABG-CoQA} & \textbf{AmbigQA-w-Ctx} & \textbf{AQuA} & \textbf{VQ-FA} & \textbf{Avg} \\
\midrule
\multirow{6}{*}{\shortstack[l]{Qwen3-4B\\(Qwen2.5-VL-3B)}}
  & Direct Prompt~\cite{kadavath2022language}        & 0.542 & 0.525 & 0.359 & 0.525 & 0.488 \\
  & EigV(Agree)~\cite{lin2024generating}              & 0.521 & 0.498 & 0.644 & 0.562 & 0.556 \\
  & EigV(Disagree)~\cite{lin2024generating}                 & 0.543 & 0.503 & 0.647 & \underline{0.587} & 0.570 \\
  & Semantic Entropy~\cite{kuhn2023semantic}          & 0.493 & 0.491 & 0.674 & 0.477 & 0.534 \\
  & Ours (Black-Box)                                  & \underline{0.615} & \textbf{0.630} & \textbf{0.755} & 0.563 & \textbf{0.641} \\
  & Ours (White-Box)                                  & \textbf{0.629} & \underline{0.530} & \underline{0.708} & \textbf{0.613} & \underline{0.620} \\
\midrule
\multirow{6}{*}{\shortstack[l]{Llama-3.1-8B\\(InternVL3-8B)}}
  & Direct Prompt~\cite{kadavath2022language}        & 0.509 & 0.545 & 0.594 & 0.477 & 0.531 \\
  & EigV(Agree)~\cite{lin2024generating}              & 0.553 & 0.552 & 0.671 & 0.575 & 0.588 \\
  & EigV(Disagree) ~\cite{lin2024generating}                 & 0.565 & 0.519 & 0.586 & 0.582 & 0.563 \\
  & Semantic Entropy~\cite{kuhn2023semantic}          & 0.560 & \underline{0.558} & 0.661 & 0.532 & 0.578 \\
  & Ours (Black-Box)                                  & \textbf{0.634} & 0.501 & \textbf{0.748} & \underline{0.603} & \underline{0.621} \\
  & Ours (White-Box)                                  & \underline{0.607} & \textbf{0.562} & \underline{0.712} & \textbf{0.636} & \textbf{0.629} \\
\midrule
\multirow{6}{*}{Gemma-3-27B}
  & Direct Prompt~\cite{kadavath2022language}        & 0.518 & 0.501 & \underline{0.719} & 0.563 & 0.575 \\
  & EigV(Agree)~\cite{lin2024generating}              & 0.532 & \underline{0.558} & 0.631 & 0.509 & 0.557 \\
  & EigV(Disagree)~\cite{lin2024generating}                 & 0.509 & 0.527 & 0.551 & 0.517 & 0.526 \\
  & Semantic Entropy~\cite{kuhn2023semantic}          & 0.506 & 0.556 & 0.593 & 0.505 & 0.540 \\
  & Ours (Black-Box)                                  & \textbf{0.650} & 0.516 & \textbf{0.790} & \underline{0.585} & \textbf{0.635} \\
  & Ours (White-Box)                                  & \underline{0.580} & \textbf{0.563} & 0.714 & \textbf{0.594} & \underline{0.613} \\
\bottomrule
\end{tabular}
\end{table*}

\subsection{Experimental Setting}
\label{ssec:setting}

\paragraph{Models and Datasets.}
The main results use three base LLMs that span three model families and a $4$B--$27$B scale range. For text QA we use Qwen3-4B-Instruct~\cite{yang2025qwen3}, Llama-3.1-8B-Instruct~\cite{grattafiori2024llama}, and Gemma-3-27B-it~\cite{Kamath2025Gemma3T}. For vision QA we use the corresponding vision-language counterparts Qwen2.5-VL-3B-Instruct~\cite{bai2025qwen3}, Gemma-3-27B-it, and InternVL3-8B~\cite{zhu2025internvl3}. All experiments are run on a single NVIDIA A100 80\,GB GPU.

We evaluate on four English QA benchmarks with explicit ambiguity annotations. ABG-CoQA~\cite{guo2021abgcoqa} is a passage-grounded conversational QA dataset with ambiguity-annotated, from which we use a subsample of 200 items balanced between ambiguous and unambiguous. AmbigQA-w-Ctx is built from the configuration of AmbigQA~\cite{min-etal-2020-ambigqa} by sampling 500 validation items, and we attach a Wikipedia passage as context for each question. AQuA~\cite{jang2026aqua} provides 720 referentially ambiguous and unambiguous visual questions over real images. VQ-FA~\cite{chen2025vqfocus} provides visually ambiguous and unambiguous questions and we sample 720 questions as well. 

\paragraph{Compared Methods.}
We compare two variants of our framework against four uncertainty-based baselines. All methods share the same base LLM and the same NLI scorer (\texttt{deberta-large-mnli}~\cite{he2020deberta}) when an agreement matrix is needed.

\emph{Direct Prompt}~\cite{kadavath2022language} asks the model in a single prompt whether the question is ambiguous, and uses $\Pr(\text{YES})$ as the score. \emph{EigV(Agree)}~\cite{lin2024generating} draws $K{=}10$ samples from the original question and reports the eigenvalue dispersion of the NLI agreement matrix. \emph{EigV(Disagree)}~\cite{lin2024generating} reuses the same samples and use NLI disagreement matrix instead. \emph{Semantic Entropy}~\cite{kuhn2023semantic,farquhar2024detecting} clusters the same samples by bidirectional NLI entailment and reports the entropy over cluster sizes. 

\paragraph{Configurations.}
Every sampling-based method shares the same decoding strategy with temperature $0.5$, top-$p$ $0.9$, $\text{max\_new\_tokens}{=}128$. Our perturbation variants draw 1 greedy sample (but less samples from the original question) from each perturbed question, with $k{=}3$ top-attributed words and up to $v{=}4$ part-of-speech-compatible WordNet variants per word. DiffMask uses 30 optimization steps with 4 greedy decode steps per step ($\lambda{=}0.5$). For Gemma-3-27B-it we load weights in bf16 with frozen parameters and gradient checkpointing, and the DiffMask attribution additionally runs through INT8 quantization via \texttt{bitsandbytes} to fit its autoregressive forward chain in memory. We report ROC-AUC (AUROC) for binary ambiguity classification on items where every method returns a valid score following previous works~\cite{lin2024generating}.

\subsection{Main Results (RQ1)}
\label{ssec:rq1}

This experiment studies whether our perturbation based detector can better tell ambiguous questions from unambiguous questions. In our method, we make small, label preserving, and semantics blind changes to the input, and then check whether the answers become less stable. We do this experiment to test our idea that applying small and semantic-blind changes to the input keeps the true ambiguity signal better than just looking at the output variance. Following the setup in Section~\ref{ssec:setting}, we evaluate on four QA benchmarks across model families and compare our two variants against four uncertainty-based baselines. We show the results in Table~\ref{tab:main}. We have the following observations:

\noindent $\bullet$ Across all three model families, at least one variant of our method achieves the best average AUROC, and our two variants together occupy the top two average positions on every model block, with an absolute improvement of up to 0.07 in average AUROC over the strongest baseline. This shows that adding label-preserving input perturbations on top of multi-sample scoring produces a more reliable ambiguity signal.

\noindent $\bullet$  The improvement is largest on the two visual ambiguity benchmarks, where our black-box variant on AQuA goes up to 0.790 AUROC. Referential and visual focus ambiguity is harder to surface from output-side variance alone because the model can return a fluent answer for any chosen referent, so the gain from our perturbation is larger here. 

\noindent $\bullet$ On the two text benchmarks, the black-box variant is generally stronger on ABG-CoQA across all three models, while the gain on AmbigQA-w-Ctx is smaller than on the other three datasets especially for white-box variant. AmbigQA-w-Ctx attaches a long Wikipedia passage to each question, so the model often has enough contextual evidence to commit to one reading regardless of how the question is perturbed. This suggests that input-side perturbation is most useful when the question itself carries the ambiguity, and that long retrieved context can partially mask the ambiguity signal by stabilizing the model's answer distribution.


\subsection{Analysis of Different Importance and Perturbations (RQ2)}
\label{ssec:rq2}

\begin{figure}
    \centering
    \includegraphics[width=0.9\linewidth]{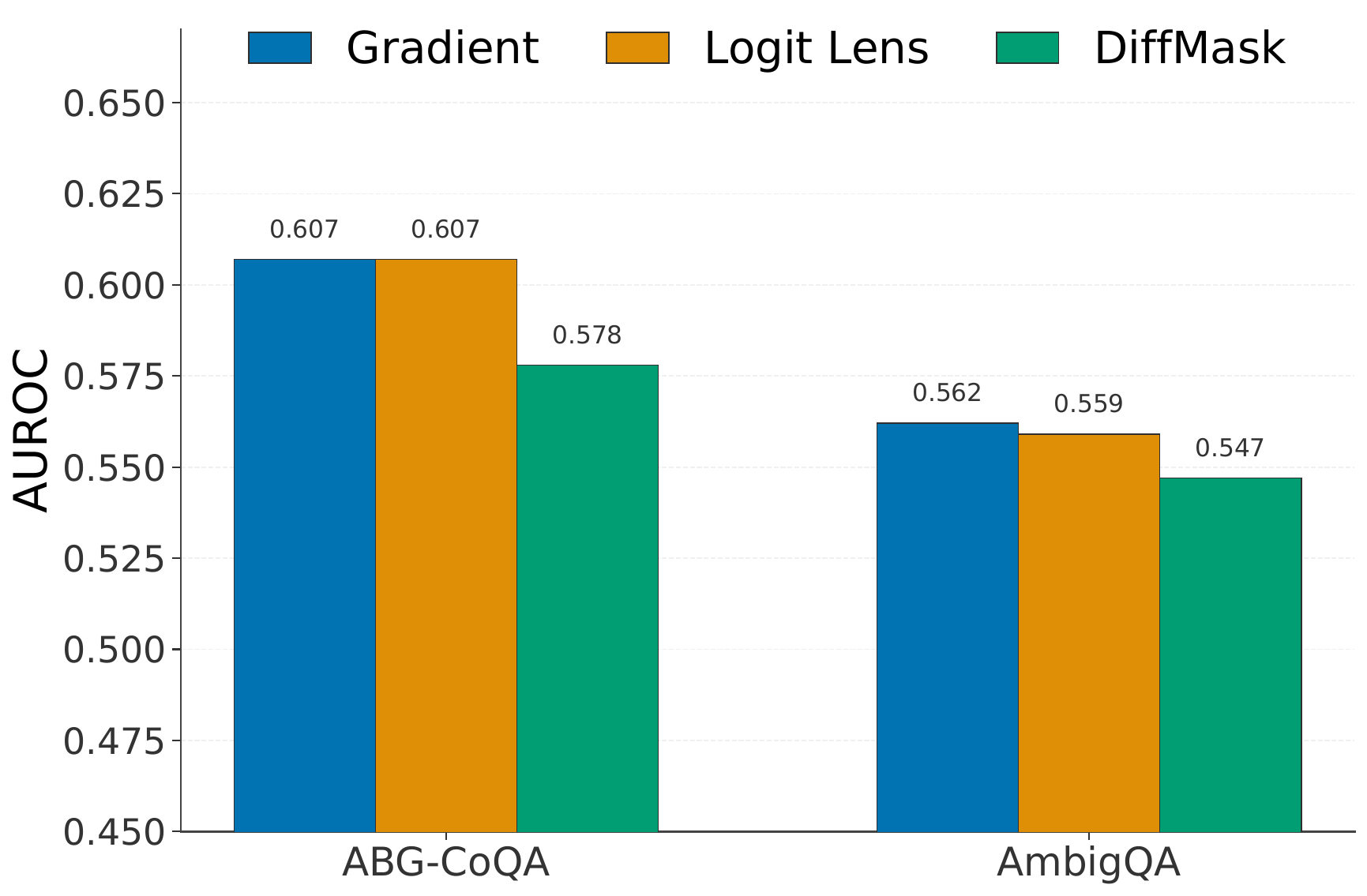}
    \caption{Performance comparing different importance methods in two datasets on Llama. The results show that though the performance of importance methods is stable, Vanilla Gradient shows a slightly better result.}
    \vspace{-2mm}
\label{fig:importance_textonly}
\end{figure}

\textbf{Choice of Importance Methods.} We test whether the importance method inside the white-box detector
matters for detection. Using Vanilla Gradient, Logit Lens, and DiffMask
(Section~\ref{ssec:whitebox_perturbation}) as three ways to select which
tokens to perturb, we report AUROC on ABG-CoQA and AmbigQA in
Figure~\ref{fig:importance_textonly}. More results can be found in Appendix~\ref{ssec:appendix_different_importance}. The results show that three methods achieve similar AUROC scores on each dataset and model, with the gap between the best and worst remaining around 0.02, and Vanilla Gradient being slightly more stable across settings. This indicates that the white-box detector does not rely on one specific importance method chosen by us, and that the ambiguity signal is mainly carried by the perturbation.



\paragraph{Choice of Different Perturbation Rules.}

\begin{figure}
    \centering
    \includegraphics[width=\linewidth]{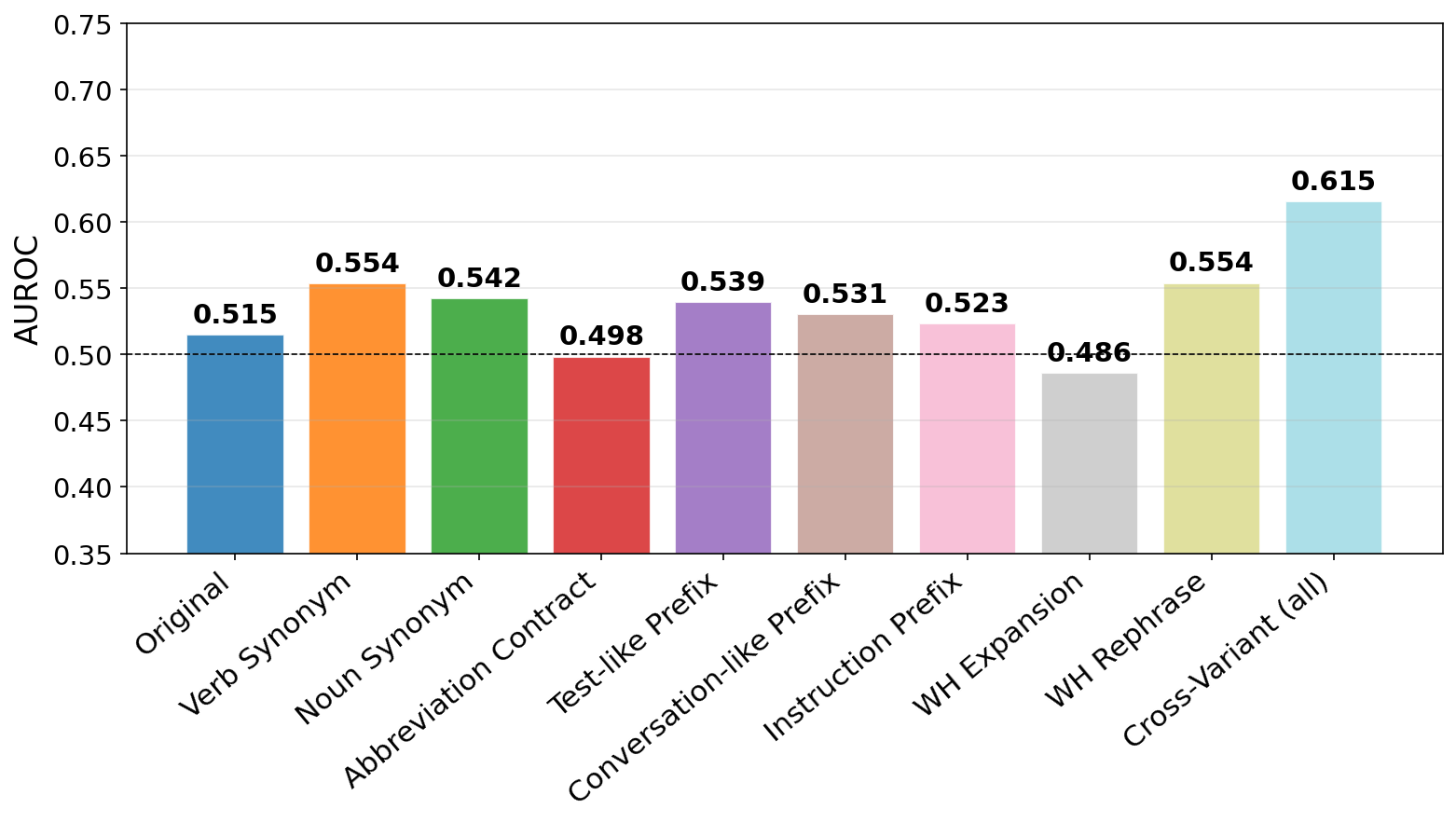}
    \caption{AUROC comparison across individual perturbation rules and their combined cross-variant setting. While several single perturbations improve over the original input, combining all perturbation variants achieves the strongest ambiguity detection performance.}
    \vspace{-2mm}
    \label{fig:different_rules}
\end{figure}

Figure~\ref{fig:different_rules} reports detection AUROC for each black-box rule individually, alongside the cross-variant setting that pools the answers from the original question and all rules into a single dispersion score. Cross-variant settings achieve the best performance. This confirms that the ambiguity signal is distributed across heterogeneous perturbations rather than concentrated in any one, and that aggregating the answer distribution over the full family of rules is what makes detection effective.

\subsection{Mitigation Performance (RQ3)}
\label{ssec:rq3}

Detection is valuable only to the extent that it improves downstream answers. We therefore evaluate whether driving mitigation with the detector's output yields higher answer correctness than generation without detection. To this end we introduce \emph{UQ-Augmented Re-asking} (UAR), which reuses the perturbation and uncertainty artifacts produced during detection and spends a single additional forward pass to recover a more reliable answer.

\paragraph{Method.}
For each question, the detector has already sampled the model under the original phrasing and every rewrite rule. UAR ranks rules by the dispersion they induce relative to the original, $\delta(r) = U(\text{rule } r) - U(\text{ORIGINAL})$, keeps the top $k$ with positive $\delta$ (default $k=3$), and for each extracts the rewritten question and its modal answer. These (rewrite, answer) pairs are inserted into the prompt as statements of the form ``when rephrased as $X$, the model answers $Y$,'' with an instruction to weigh the passage and commit to a single answer. A question for which no rule has a positive $\delta$ gets the unchanged baseline prompt, so re-asking fires only when the detector has localized interpretation-sensitive behavior.

\paragraph{Setup and results.}
We compare against \textit{Original}, the model's first answer to the question with no uncertainty information; both conditions share the same model, context, and decoding setup (greedy, single sample), so any difference comes from the prompt. As Table~\ref{tab:uar} shows, UAR improves Exact Match on every subset of both datasets, by $+8.15$ to $+18.07$ points. The gain also appears on the gold non-ambiguous subset because re-asking fires on questions whose \emph{model} answers disagree across rewrites, a broader signal than the human label: many gold non-ambiguous questions still destabilize the model, and the injected evidence helps it commit to a single answer. Additional controlled comparisons with generic re-prompting and self-sampling are provided in Appendix~\ref{app:uar_controls}.

\begin{table}[t]
\centering
\small
\caption{UAR vs.\ the original baseline on
Qwen3-4B, greedy decoding. UAR re-prompts the model with detector-derived evidence; Original is the model's first answer to the unperturbed question.}
\resizebox{\columnwidth}{!}{%
\begin{tabular}{llrrr}
\toprule
Config & Group  & Original & UAR & $\Delta$acc \\
\midrule
\multirow{2}{*}{AmbigQA-w-Ctx}
  & ambiguous      & 12.19 & \textbf{26.52} & $+14.33$ \\
  & non\_ambig.    & 16.28 & \textbf{34.35} & $+18.07$ \\
\midrule
\multirow{2}{*}{ABG-CoQA}
  & ambiguous      & 22.76 & \textbf{34.15} & $+11.39$ \\
  & non\_ambig.    & 17.06 & \textbf{25.21} & $+8.15$ \\
\bottomrule
\end{tabular}%
}
\label{tab:uar}
\end{table}

\subsection{Ablation Study}

\begin{figure}
    \centering
    \includegraphics[width=0.9\linewidth]{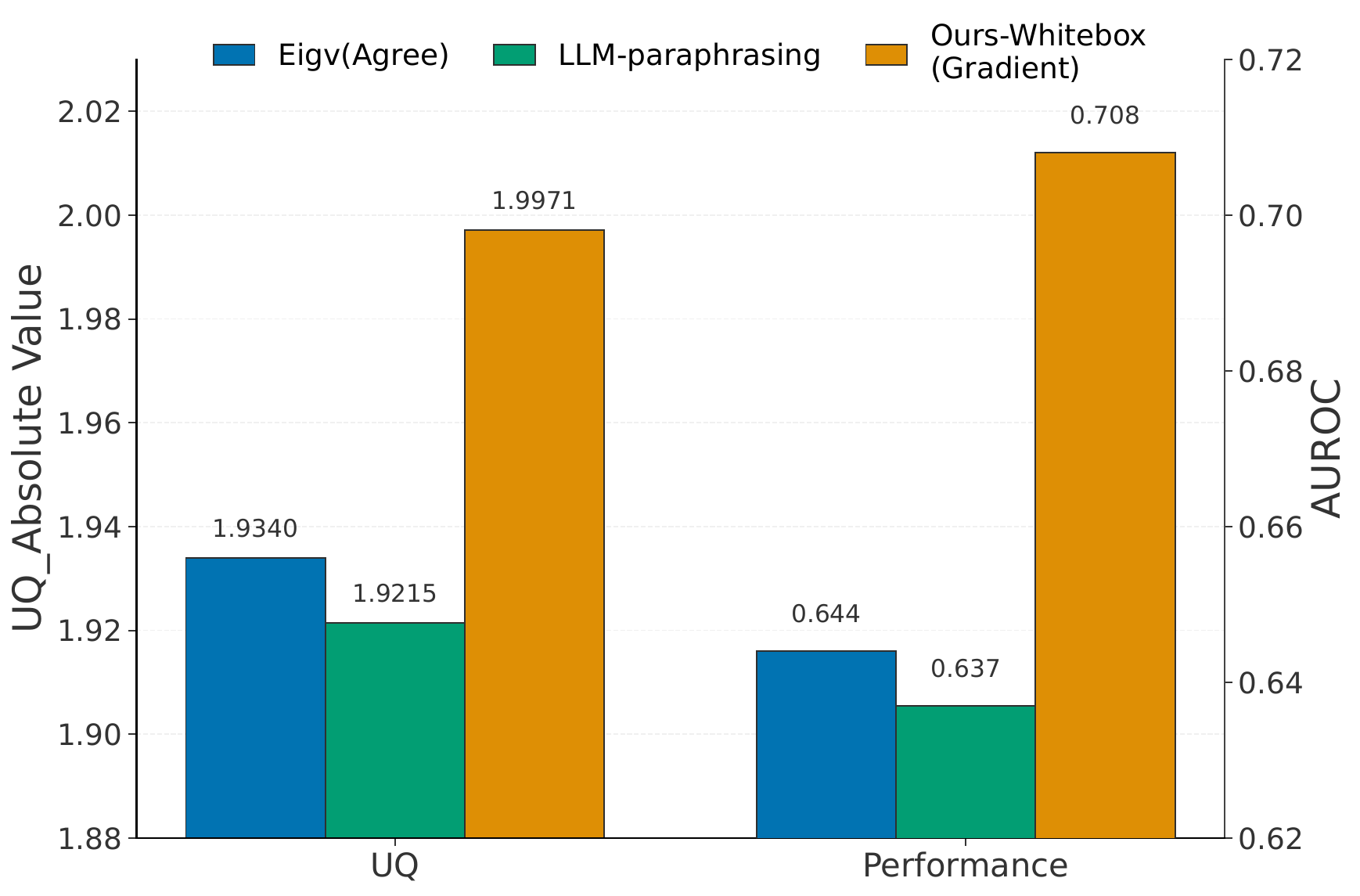}
    \caption{Comparison of performance and absolute value of uncertainty on ambiguous questions from AquA. The results show that directly using LLM-paraphrasing can actually reduce the uncertainty.}
    \label{fig:uq_absotule}
\end{figure}

\paragraph{Influence of Semantic-blind Perturbation}
This experiment tests whether the semantic-blind constraint on the perturbation step is actually necessary. As discussed in Section~\ref{ssec:overall_framework}, our framework requires the perturbation to be low-capacity and semantics-blind, so that the input is rewritten without choosing among its possible readings. A natural alternative is to let an LLM directly paraphrase the question, which is a high-capacity transformation that may resolve the ambiguity during rewriting. We report both the absolute uncertainty value and the AUROC in Figure~\ref{fig:uq_absotule} using Qwen2.5-VL-3B and AquA. The results show that replacing our semantic-blind perturbation with LLM paraphrasing lowers the absolute uncertainty value to even below the EigV(Agree) baseline that uses no input perturbation, and the AUROC also drops from 0.708 to 0.637. This shows that LLM paraphrasing does not just fail to add a useful signal, it can potentially remove ambiguity that the baseline could still pick up from sampling alone.

\paragraph{Rule Execution: Ours vs.\ LLM-based.}

In the black-box setting, a fixed rule can be executed either deterministically, with WordNet synonyms and string templates, and the rewrite is strictly semantics-blind, or by handing the same rule to an LLM as an instruction. Unlike free LLM paraphrasing, the rule is fixed in advance; the question is whether this constraint is enough to stop the LLM from understanding, and thereby disambiguating, the original question. The two realizations share an identical rule inventory and differ only in the capacity of the mechanism that executes them. As shown in Table~\ref{tab:executor}, across four dataset settings spanning text (Qwen3-4B) and vision (Qwen2.5-VL-3B), ours perform better than the LLM-based one. It shows that even a tightly constrained rule, once routed through a model that must understand the input, erodes rather than preserves the ambiguity signal, which supports our use of semantics-blind rewrites for detection.

\paragraph{Semantic Consistency of Perturbations.}
To verify that our perturbations remain close to the original inputs, we compute the cosine similarity between original and perturbed inputs using Qwen3-Embedding-0.6B and Qwen3-Embedding-4B. For text QA, we compare the full textual context; for visual QA, we compare only the edited question since the image is unchanged. As shown in Table~\ref{tab:semantic_consistency}, the average cosine similarities range from 0.883 to 0.999 across all datasets and model families, indicating that our perturbations introduce only limited semantic drift and supporting the intended label-preserving design.

\begin{table}[t]
\centering
\small
\begin{tabular}{llcc}
\toprule
Model & Dataset & 0.6B & 4B \\
\midrule
\multirow{4}{*}{Qwen}
& ABG-CoQA      & 0.9960 & 0.9989 \\
& AmbigQA-w-Ctx & 0.9781 & 0.9842 \\
& AQuA          & 0.9107 & 0.9191 \\
& VQ-FA         & 0.8843 & 0.8923 \\
\midrule
\multirow{4}{*}{Llama/InternVL}
& ABG-CoQA      & 0.9952 & 0.9985 \\
& AmbigQA-w-Ctx & 0.9749 & 0.9835 \\
& AQuA          & 0.9037 & 0.9097 \\
& VQ-FA         & 0.8834 & 0.8903 \\
\bottomrule
\end{tabular}
\caption{Cosine similarity between original and perturbed inputs using Qwen3 embedding models.}
\label{tab:semantic_consistency}
\end{table}

\begin{table}[t]
\centering
\small
\caption{Detection AUROC for two realizations of the same rule set. Our perturbations outperform LLM-based ones, while remaining semantics-blind and cheaper.}
\resizebox{\columnwidth}{!}{%
\begin{tabular}{llcc}
\toprule
Model & Dataset & Ours & LLM \\
\midrule
\multirow{2}{*}{Qwen3-4B}
  & ABG-CoQA & $\textbf{0.615}$ & $0.598$ \\
  & AmbigQA  & $\textbf{0.630}$ & $0.617$ \\
\midrule
\multirow{2}{*}{Qwen2.5-VL-3B}
  & AQuA & $\textbf{0.755}$ & $0.715$ \\
  & VQ-FA & $\textbf{0.563}$ & $0.538$ \\
\bottomrule
\end{tabular}%
}
\label{tab:executor}
\end{table}

\subsection{Failure Case Analysis}
\label{sec:failure_analysis}

To better understand the limitations of our detector, we analyze its errors on AQuA with Gemma-3-27B using the black-box setting. At the Youden-optimal threshold, the detector produces 376 true positives, 164 false negatives, 46 false positives, and 134 true negatives. Most false negatives are referential or deictic questions: 154 of the 164 missed ambiguous examples fall into this category, and 69 of them produce at most two distinct answers across perturbations. This suggests that the model can sometimes silently commit to one salient referent, making the ambiguity difficult to expose through surface perturbations. Among the 46 false positives, 22 are counting questions and many others require fine-grained visual perception, indicating that answer instability can also arise from model or perceptual uncertainty rather than input ambiguity.

\section{Conclusion}

This work studies ambiguity detection as a prerequisite for reliable question answering. We propose a perturbation-based framework that preserves ambiguity signals by avoiding semantic rephrasing, and instantiate it with both white-box and black-box detectors. Experiments across open-domain, conversational, and visual QA datasets show that our method more effectively distinguishes ambiguous from unambiguous questions than output-side sampling and paraphrase-based baselines. When its output drives evidence-augmented re-asking, the detector further improves downstream answer accuracy on under-specified questions.
\newpage
\section*{Limitations}

First, the black-box detector relies on a manually designed rule library built for English, so applying it to other languages requires reconstructing these rules. Second, the white-box detector depends on access to model internals, which is unavailable in many deployment settings such as closed-source APIs.

\section*{Acknowledgments}
The work was partially supported by Toyota InfoTech Labs and NSF awards \#2442477, \#2550203, \#2616632, \#2623317, \#2536297, and \#2613637. The views and conclusions in this paper should not be interpreted as representing any funding agencies.


\bibliography{custom}

\clearpage

\afterpage{\begin{figure*}[!ht]
\centering
\vspace{-3mm}
\begin{subfigure}[t]{0.32\linewidth}
  \centering
  \includegraphics[width=\linewidth]{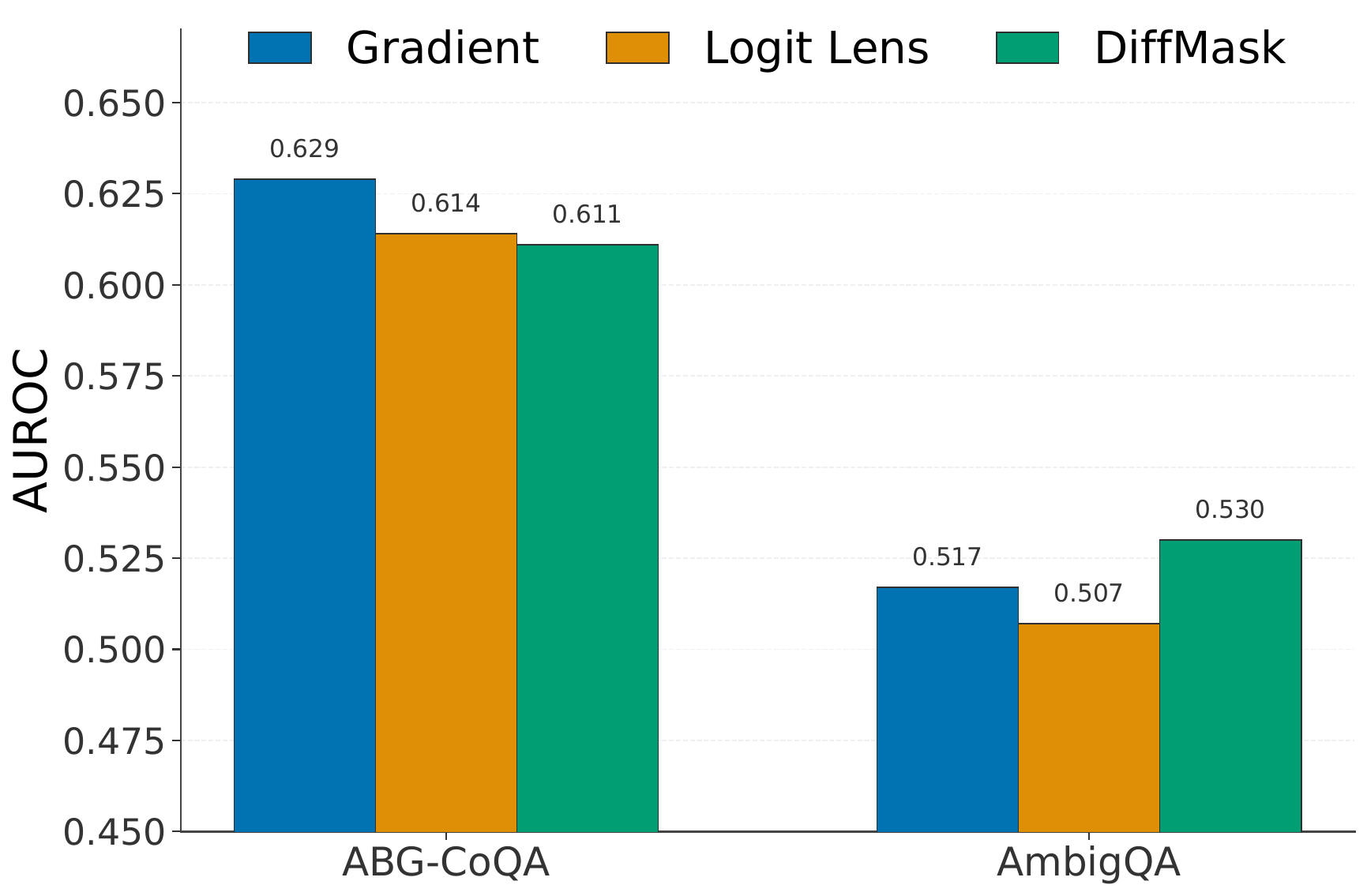}
  \vspace{-6mm}
  \caption{\small Qwen3.}
  \label{fig:sft-sensitivity:a}
\end{subfigure}\hfill
\begin{subfigure}[t]{0.32\linewidth}
  \centering
  \includegraphics[width=\linewidth]{latex/Figures/Llama_3_1_8B_eigv_entail.pdf}
  \vspace{-6mm}
  \caption{\small Llama3.1.}
  \label{fig:sft-sensitivity:b}
\end{subfigure}\hfill
\begin{subfigure}[t]{0.32\linewidth}
  \centering
  \includegraphics[width=\linewidth]{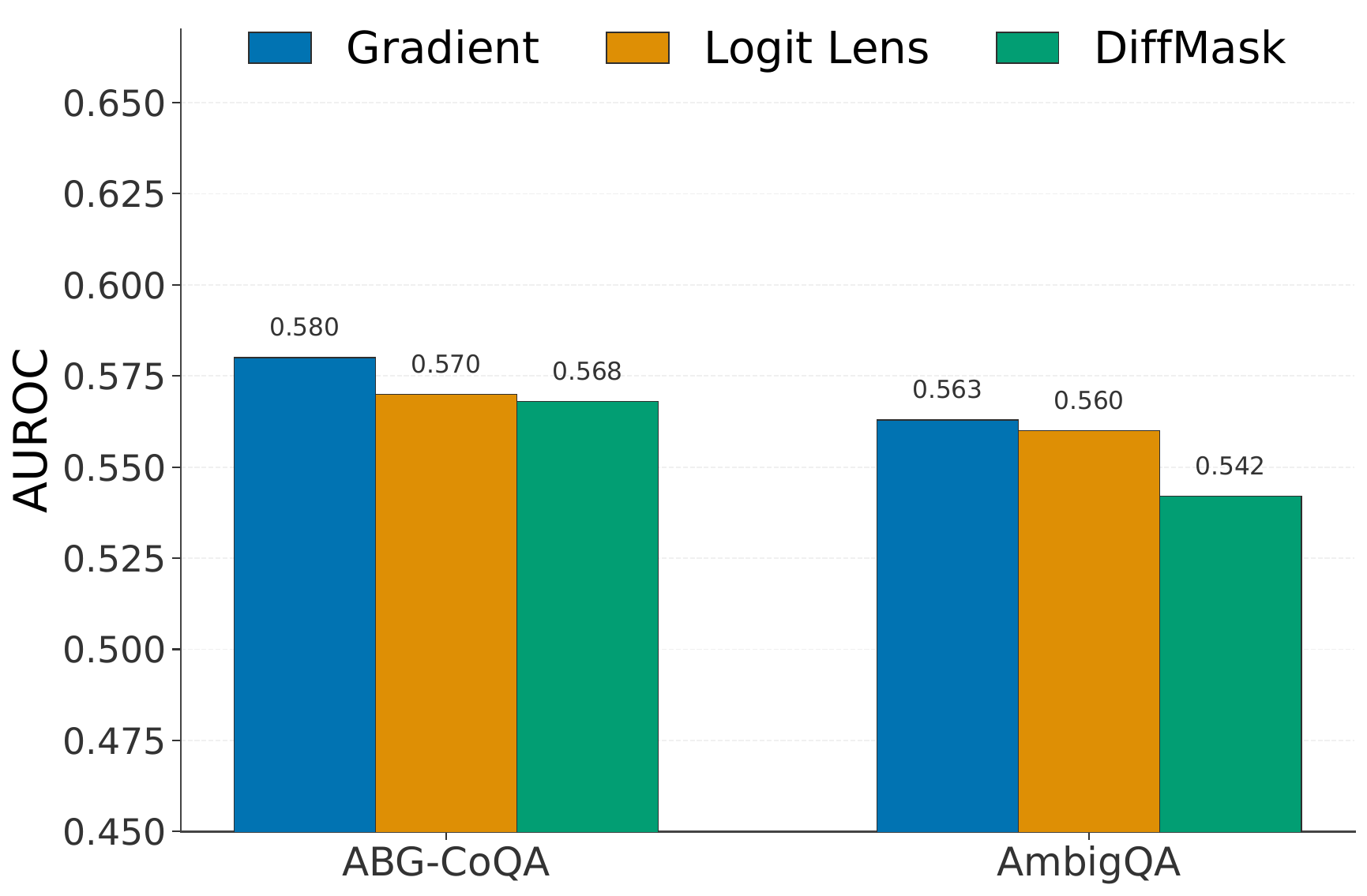}
  \vspace{-6mm}
  \caption{\small Gemma3-27b.}
  \label{fig:sft-sensitivity:c}
\end{subfigure}
\caption{Performance comparing different white-box importance methods in text-only datasets. The results show that though the performance of importance methods is stable, Vanilla Gradient shows a slightly better result.}
\label{fig:importance_textonly_full}
\end{figure*}}

\appendix

\section{Detail Technology of Uncertainty Estimation}
\label{sec:appendx_tech_uq}
We first use an NLI model to compute a pairwise semantic agreement matrix \(A\) over answers in \(R_X^{\mathcal{T}}\). For answers \(r_i\) and \(r_j\), we set
\begin{equation}
A_{ij}=\frac{1}{2}\left(p_{\mathrm{ent}}(r_i\Rightarrow r_j)+p_{\mathrm{ent}}(r_j\Rightarrow r_i)\right),
\end{equation}
where \(p_{\mathrm{ent}}\) is the entailment probability assigned by the NLI model, with \(A_{ii}=1\). We then build the degree matrix \(D\), where \(D_{ii}=\sum_j A_{ij}\), and compute the symmetric normalized graph Laplacian
\begin{equation}
L=I-D^{-1/2}AD^{-1/2}.
\end{equation}
Let \(\lambda_1,\ldots,\lambda_N\) be the eigenvalues of \(L\), where \(N=|R_X^{\mathcal{T}}|\). Following the spectral clustering motivation in Generating with Confidence, smaller eigenvalues indicate semantic clusters among sampled answers. We compute
\begin{equation}
U(X)=\mathrm{EigV}(\mathrm{Agree}(R_X^{\mathcal{T}}))
=\sum_{q=1}^{N}\max(0,1-\lambda_q).
\end{equation}
A larger value indicates that the combined original and perturbed samples separate into more semantic modes, which is the signal we use for ambiguity.

\section{More Experiments}

\subsection{More Results on Different White-box Importance}
\label{ssec:appendix_different_importance}

We test whether the importance method inside the white-box detector matters for detection using all models we used for text-only datasets and the full results can be found at Figure~\ref{fig:importance_textonly_full}. The conclusion is consistent with the main text that our choice of importance is stable and Vanilla Gradient is slightly better.

\begin{figure}[h!]
    \centering
    \includegraphics[width=0.9\linewidth]{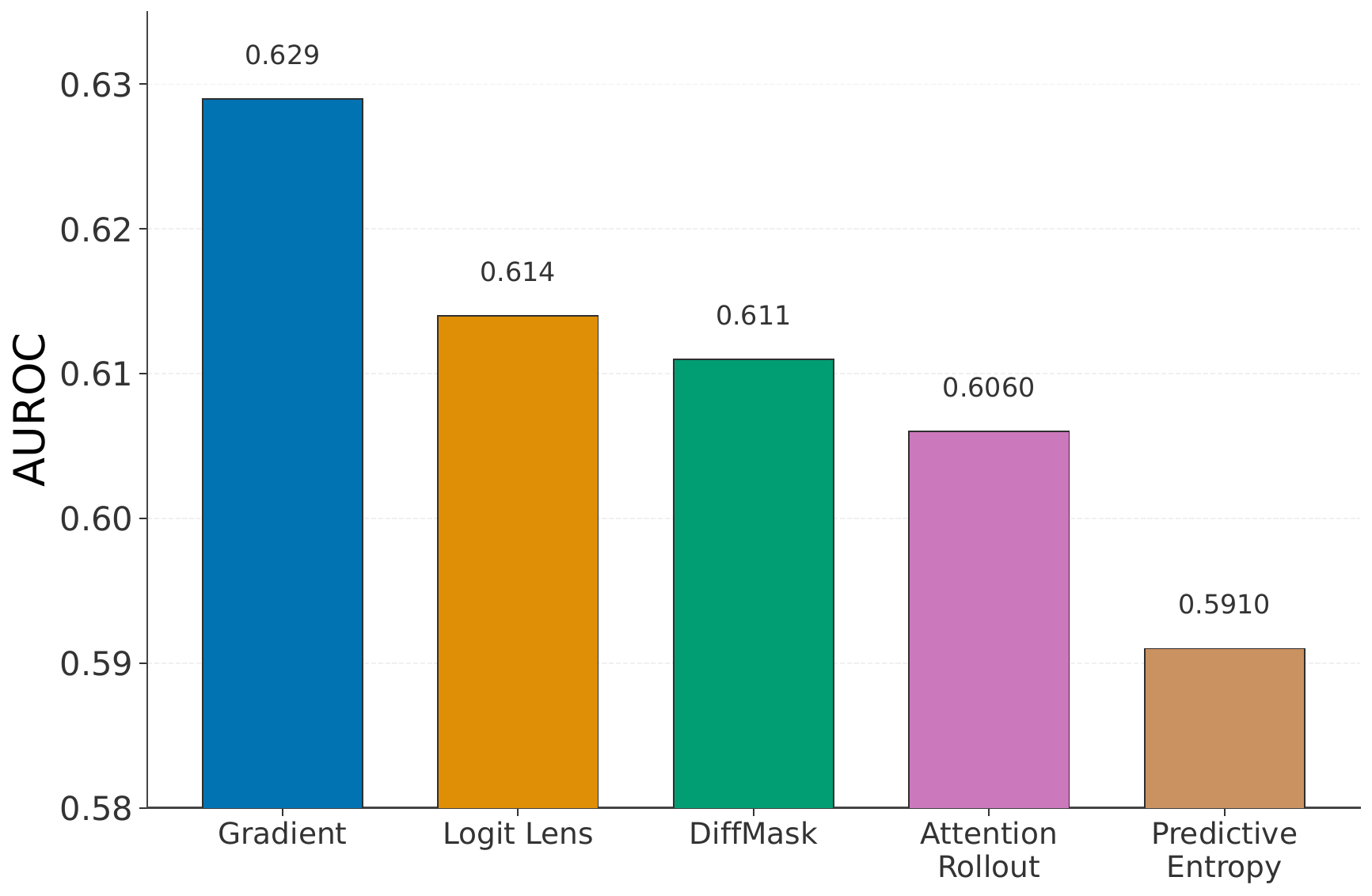}
    \caption{Performance comparing more white-box importance methods on Abg-CoQA and Qwen3. The results show that our choice is better than simple attention or predictive entropy.}
    \label{fig:white_box_ablation}
    \vspace{-5mm}
\end{figure}

\subsection{Comparison with Ambiguity-Aware Prompting}
\label{app:procot}

Most baselines in the main experiments quantify uncertainty
from sampled outputs. We additionally compare with Proactive
Chain-of-Thought (ProCoT) prompting~\citep{deng-etal-2023-prompting},
an ambiguity-aware prompting approach that explicitly asks the
model to determine whether a question is ambiguous before
answering or requesting clarification.

Table~\ref{tab:procot} reports the mean AUROC across the four
datasets for each model family. ProCoT substantially improves
over directly asking the model whether the question is ambiguous,
showing that explicit ambiguity-aware prompting is a stronger
baseline than a direct prompt. Nevertheless, both variants of our
perturbation-based detector achieve higher average AUROC. This
supports our central motivation that ambiguity can be detected more
reliably by measuring response instability under constrained,
semantics-blind perturbations than by asking the model to reason
about the ambiguity directly.

\begin{table}[t]
\centering
\small
\begin{tabular}{lcc}
\toprule
Method & Qwen & Llama/InternVL \\
\midrule
Direct Prompt      & 0.488 & 0.531 \\
ProCoT             & 0.574 & 0.585 \\
Ours (Black-Box)   & \textbf{0.641} & 0.621 \\
Ours (White-Box)   & 0.620 & \textbf{0.629} \\
\bottomrule
\end{tabular}
\caption{Mean ambiguity-detection AUROC across the four datasets.
ProCoT provides a stronger ambiguity-aware prompting baseline than
Direct Prompt, while our perturbation-based detectors achieve higher
average AUROC for both model families.}
\label{tab:procot}
\end{table}

\subsection{Controlled Analysis of UQ-Augmented Re-asking}
\label{app:uar_controls}

To test whether the gain of UAR comes from ambiguity-aware evidence rather than simply re-prompting with additional context, we compare against two controls on Qwen3-4B. \textbf{Hint} uses a generic re-prompt indicating that the answer may be uncertain, while \textbf{SelfSample} uses the same evidence budget as UAR but replaces detector-selected evidence with additional answers sampled from the original question. As shown in Table~\ref{tab:uar_controls}, UAR consistently outperforms both controls, indicating that detector-derived evidence provides benefits beyond generic re-prompting or additional sampled context.

\begin{table}[t]
\centering
\small
\resizebox{\columnwidth}{!}{
\begin{tabular}{llrrrr}
\toprule
Dataset & Group & Original & Hint & SelfSample & UAR \\
\midrule
AmbigQA-w-Ctx
& ambiguous     & 12.19 & 12.36 & 20.92 & \textbf{26.52} \\
& non-ambig.    & 16.28 & 16.79 & 25.19 & \textbf{34.35} \\
\midrule
ABG-CoQA
& ambiguous     & 22.76 & 24.39 & 28.46 & \textbf{34.15} \\
& non-ambig.    & 17.06 & 17.92 & 20.28 & \textbf{25.21} \\
\bottomrule
\end{tabular}
}
\caption{Exact Match comparison between UAR and controlled re-prompting baselines on Qwen3-4B.}
\label{tab:uar_controls}
\end{table}

\subsection{Sensitivity to Perturbation Budget}
\label{app:perturbation_budget}

We further study whether the black-box detector depends on the full perturbation library by varying the number of available rules on AmbigQA with Qwen3-4B. As shown in Table~\ref{tab:perturbation_budget}, performance generally improves as more perturbation rules are included and saturates around nine rules, suggesting that the detector benefits from broader perturbation coverage rather than relying on a single specific rule.

\begin{table}[t]
\centering
\resizebox{\columnwidth}{!}{
\begin{tabular}{lrrrrrr}
\toprule
\# Rules & 1 & 3 & 5 & 7 & 9 & 11 \\
\midrule
AUROC & 0.587 & 0.607 & 0.600 & 0.618 & 0.630 & 0.630 \\
\bottomrule
\end{tabular}}
\caption{Ambiguity-detection AUROC under different black-box perturbation budgets on AmbigQA with Qwen3-4B.}
\label{tab:perturbation_budget}
\end{table}

\subsection{More Ablation}
\paragraph{More White-box Importance}

To test whether our choice of white-box importance makes sense, we compare with our white-box methods with more white-box importance methods:

\noindent $\bullet$ attention: recursively multiplies per-layer attention matrices to compute the total
information flow from each input token to the output. \\
\noindent $\bullet$ predictive entropy: assigns each input token an importance score equal to the entropy of the model's next-token distribution at that position.

We show the results in Figure~\ref{fig:white_box_ablation}, which shows that simply using attention or entropy cannot get better results, and our choice of importance methods is reasonable.


\section{Detail Methods for Black-Box Perturbation}
\label{app:blackbox_details}

This appendix lists the full set of rules used by the black-box perturbation module described in Section~\ref{ssec:blackbox_perturbation}. All rules are semantics-blind, meaning that each rule fires on surface features such as part-of-speech tag, orthographic form, or the leading wh word, and never consults a model-generated reading of the input. The rules are organized into two groups according to the granularity of the edit.

\subsection{Group A: Single Word Level}

Rules A1 to A4 substitute a single content word with a WordNet~\cite{miller1995wordnet} synonym drawn from the first synset of the target word, with multi-word, very short, length-mismatched, and rare candidates filtered out. The selected synonym is then reinflected to match the tense, number, or degree of the original word.

\noindent $\bullet$ \textbf{A1: Verb Synonym.} Replace a verb with a WordNet synonym, reinflected to match the original verb form. Example: \emph{Who \underline{won} the game?} $\rightarrow$ \emph{Who \underline{prevailed} the game?}

\noindent $\bullet$ \textbf{A2: Noun Synonym.} Replace a noun with a WordNet synonym, with number preserved. Example: \emph{When was the \underline{building} completed?} $\rightarrow$ \emph{When was the \underline{structure} completed?}

\noindent $\bullet$ \textbf{A3: Adjective Synonym.} Replace an adjective with a WordNet synonym, with degree preserved. Example: \emph{What is the \underline{largest} city?} $\rightarrow$ \emph{What is the \underline{biggest} city?}

\noindent $\bullet$ \textbf{A4: Adverb Synonym.} Replace an adverb with a WordNet synonym. Example: \emph{How \underline{quickly} does it grow?} $\rightarrow$ \emph{How \underline{rapidly} does it grow?}

\noindent $\bullet$ \textbf{A5: Abbreviation Expand.} Expand a contraction into its full form. Example: \emph{\underline{Who's} the author?} $\rightarrow$ \emph{\underline{Who is} the author?}

\noindent $\bullet$ \textbf{A6: Abbreviation Contract.} The inverse of A5, contracting a full function-word sequence into its abbreviated form. Example: \emph{\underline{Who is} the author?} $\rightarrow$ \emph{\underline{Who's} the author?}

\subsection{Group B: Sentence and Phrase Level}

Rules B1 to B3 keep the question stem verbatim and only change its leading register through a short prefix. Rules B4 and B5 act on the wh word at the beginning of the question through a small fixed mapping table, and both are guarded so that they do not fire when the wh word is acting as a determiner, as in \emph{What color was it}.

\noindent $\bullet$ \textbf{B1: Test-like Prefix.} Prepend a short test-style phrase. Examples include \emph{According to the passage}, \emph{Based on the text above}, and \emph{As stated in the document}.

\noindent $\bullet$ \textbf{B2: Conversation-like Prefix.} Prepend a short conversational opener. Examples include \emph{Hey so}, \emph{Quick question}, and \emph{By the way}.

\noindent $\bullet$ \textbf{B3: Instruction Prefix.} Prepend a short instruction-style phrase. Examples include \emph{Please answer the following question}, \emph{Your task is to answer}, and \emph{Respond to this question}.

\noindent $\bullet$ \textbf{B4: WH Expansion.} Replace a bare wh word with an equivalent determiner phrase. Example: \emph{\underline{Who} won?} $\rightarrow$ \emph{\underline{Which person} won?}

\noindent $\bullet$ \textbf{B5: WH Rephrase.} Replace the wh word with a semantically aligned alternative. Example: \emph{\underline{How} did it happen?} $\rightarrow$ \emph{\underline{By what means} did it happen?}

\end{document}